\documentclass[letterpaper]{article} 
\usepackage[submission]{aaai24}  
\usepackage{times}  
\usepackage{helvet}  
\usepackage{courier}  
\usepackage[hyphens]{url}  
\usepackage{graphicx} 
\usepackage{multirow}
\usepackage[inline]{enumitem}
\usepackage{natbib}  
\usepackage{caption} 
\usepackage{algorithm}
\usepackage{algorithmic}

\usepackage{newfloat}
\usepackage{listings}
\DeclareCaptionStyle{ruled}{labelfont=normalfont,labelsep=colon,strut=off} 
\floatstyle{ruled}
\newfloat{listing}{tb}{lst}{}
\floatname{listing}{Listing}
\title{Source Prompt: Coordinated Pre-training of Language Models on Diverse Corpora from Multiple Sources}

\author{San Zhang}

\author{
    Yipei Xu, Dakuan Lu, Jiaqing Liang, Xintao Wang, Yipeng Geng, Yingsi Xin, Hengkui Wu, Ken Chen, ruiji zhang, Yanghua Xiao
}
\affiliations{
    Fudan University, SuperSymmetry Technologies
}

\usepackage{bibentry}

\begin{document}

\maketitle

\begin{abstract}
Pre-trained language models (PLMs) have established the new paradigm in the field of NLP. 
For more powerful PLMs, one of the most popular and successful way is to continuously scale up sizes of the models and the pre-training corpora.
These large corpora are generally obtained by converging smaller ones from multiple sources, they are thus growing increasingly diverse. 
However, the side-effects of these colossal converged corpora remain understudied. 
In this paper, we identify the disadvantage of heterogeneous corpora from multiple sources for pre-training PLMs.
Towards coordinated pre-training on diverse corpora, we further propose source prompts (SP), which explicitly prompt the model of the data source at the pre-training and fine-tuning stages. Results of extensive experiments demonstrate that PLMs pre-trained with SP on diverse corpora gain significant improvement in various downstream tasks.
\end{abstract}

\section{Introduction}

Recently, pre-trained language models (PLMs) have markedly enhanced state-of-the-art performance in natural language processing (NLP). They introduced a new approach using pre-training followed by fine-tuning. Specifically, these models glean extensive linguistic knowledge from unsupervised pre-training on vast corpora. To enhance the capabilities of PLMs, the most effective practice has been found to involve developing larger models pre-trained on enormous, varied corpora~\cite{2019t5,brown2020language,wu2021yuan}.

Corpus expansion is thus crucial for pre-training large PLMs. This is typically achieved by combining several corpora~\cite{gao2020pile,devlin2018bert,yang2020finbert} from various sources, including large Internet corpora collected using common crawlers~\cite{2019t5,xue2020mt5,yuan2021wudaocorpora,wu2021yuan}. Consequently, many diverse corpora are used in training the PLM, ensuring adaptability for numerous downstream tasks. Table~\ref{tab:typicalplm-corpus} presents a selection of popular general or domain-specific PLMs, each pre-trained from varied corpora.


Increasing corpus size by integrating more heterogeneous corpora does not always enhance PLMs' performance. For some downstream tasks or datasets, pre-training on unrelated sub-corpora may be harmful. Table~\ref{table:wikit5} illustrates this. A T5 model pre-trained on a combined Wikipedia and Toronto Books Corpus (TBC)~\cite{zhu2015aligning}, totalling 20 GB, achieves a SuperGLUE score of 73.24. However, the same model pre-trained on the much larger but more heterogeneous 745 GB C4~\cite{2019t5} corpus gets a lower score of 71.36. The C4 corpus is larger and more varied, but its quality is worse than the Wikipedia and TBC corpuses. Therefore, the varied distribution of large corpora challenges the performance of large PLMs in some benchmarks.


\begin{table*}[!htb]
\centering
\resizebox{2\columnwidth}{!}{
\begin{tabular}{l l l}
\hline
\textbf{PLM} & \textbf{Corpus} & \textbf{Sources}\\
\hline
BERT
& Corpus of BERT
& BooksCorpus (TBC) and English Wikipedia 
\\
GPT2-chinese~\cite{zhao2019uer}
& CLUECorpussSmall
& News, Social networking sites, Chinese Wikipedia and Reviews
\\
GPT-3
& Corpora of GPT-3
& Common Crawl (filtered), WebText2, Books1, Books2, Wikipedia
\\
CPM-2~\cite{zhang2021cpm}
& WuDaoCorpora
& Encyclopedia, Novels, QA, Scientific Literature, E-book, News, and Reviews.
\\
BioBERT~\cite{lee2020biobert}
& Biomedical corpora
& PubMed abstracts and PMC full-text articles
\\
FinBERT~\cite{yang2020finbert}
& Financial corpora
& Corporate Reports, Earnings Call Transcripts, Analyst Reports
\\
BBT-FinT5\footnote{https://github.com/ssymmetry/BBT-FinCUGE-Application}
& BBT-FinCorpus
& Corporate Reports, Analyst Reports, Stock Bar Forum and Financial News
\\
\hline
\end{tabular}
}
\caption{Pre-training corpora and corresponding diverse sources of several typical PLMs.}
\label{tab:typicalplm-corpus}
\end{table*}

\begin{table}[!tb]
\centering
\begin{tabular}{l l c c}
\hline
\textbf{Corpus} & \textbf{Size} & \textbf{GLUE} & \textbf{SuperGLUE} \\
\hline
Wiki + TBC
& 20GB
& \textbf{83.65}
& \textbf{73.24}
\\
C4
& \textbf{745GB}
& 83.28
& 71.36
\\
\hline
\end{tabular}
\caption{Comparison between T5 pre-trained on different corpus. 
While C4 significantly outstrip Wiki+TBC in terms of scale and diversity, T5 pre-trained on C4 performs worse than the latter.
}
\label{table:wikit5}
\end{table}



To further enhance the performance of large pre-trained language models (PLMs), a critical element is effectively coordinating high data source diversity corpora. Numerous studies~\cite{wang-etal-2018-denoising,silva-etal-2018-extracting,aharoni-goldberg-2020-unsupervised,iter2021trade} emphasize the importance of data diversity in machine learning tasks. These studies propose extracting training examples that are closely similar to the downstream task to enhance performance. However, data resampling for pre-training corpora is impractical due to two main reasons:
\begin{enumerate*}[label=\alph*).]
    \item Data resampling renders unselected data unused during pre-training, thereby diminishing the utilization of corpora data. 
    \item As PLMs are designed to handle a wide range of downstream tasks, resampling for specific tasks damages the generality of PLMs.
    \item The proportion allocation of pre-training data for large-scale models is key and challenging, usually relying on prior experience. Data resampling will have an impact on the original data distribution.
\end{enumerate*}

We introduce the concept of source prompt (SP) to improve the performance of PLMs that are trained from diverse, huge, and imbalanced corpora.
Instead of using careful resampling strategies or probing good corpus proportions, we propose that corpus source serves as a clear indicator of the heterogeneous data distribution of extensive corpora. 
The source prompt is used in conjunction with the input sequence to indicate the source of the data, implemented during both pre-training and downstream tasks. 
For instance, when pre-training a model such as BERT on Wikipedia and BookCorpus concurrently, we can insert the prompts '[WIKI]' and '[BOOK]' before the input from Wikipedia and BookCorpus respectively. For a given downstream task dataset, we assign one of the pre-training corpus sources and insert the SP corresponding. This source can be either manually or automatically assigned.
Our method warrants emphasis due to four principal advantages:
\begin{enumerate}
    \item \textbf{Generalizability.} It can be directly implemented on pre-trained corpora. Its operation is entirely dependant on the sources of pre-training data, not necessitating any consideration of downstream task features. 
    \item \textbf{Data Utilization.} It ensures that all corpora can be fully utilized, eliminating the requirement for tricky data resampling.
    \item \textbf{Simplicity.} Our methodology doesn't need preprocessing and additional modules,  conserving significant computational resources.
    \item \textbf{Applicability.} Our approach and the pre-trained models can be effortlessly integrated into existing frameworks, foregoing any alternations to the model structure and training procedures.
\end{enumerate}
Our experiments reveal that our approach yields superior PLMs post pre-training on varied corpora. These models significantly outperform on diverse downstream tasks.


Our contributions are summarized as follows:
\begin{itemize}
    \item We propose the Source Prompt mechanism, designed to leverage source diversity for enhanced performance of PLMs in downstream tasks.
    \item We realize the Source Prompt concept across three ubiquitous Transformer architectures: Encoder-Only, Encoder-Decoder, and Decoder-Only. 
    \item We perform experiments on multiple pre-training corpora and various downstream datasets to verify the effectiveness of SP. The results show that PLMs pre-trained with SP on different corpora achieve significant improvements in different settings.
\end{itemize}

\section{Related Work}
In this section, we introduce two lines of related work, including efforts on data diversity in  single-task settings, and prompt-tuning methods for PLM-based multi-task learning.

\subsection{Data Diversity in Single-Task Settings}
\subsubsection{Data Selection}
Data selection is frequently employed to address issues arising from data diversity in specific tasks~\cite{silva-etal-2018-extracting, wang-etal-2018-denoising, aharoni-goldberg-2020-unsupervised}.
This technique involves choosing training samples based on their resemblance to the validation set or a reliable in-domain dataset.

For instance,
\citet{aharoni-goldberg-2020-unsupervised} employs distance-based retrieval using sentence embeddings produced by PLMs to select in-domain data.
\citet{van-der-wees-etal-2017-dynamic} presents dynamic data selection to enhance neural machine translation (NMT) and introduces gradual fine-tuning, which surpasses traditional static data selection.
\citet{wang-etal-2018-denoising} extends techniques to assess and select data for domain NMT and adapts them for denoising NMT training, using reliable data and an online data selection-based denoising curriculum.

However, while data selection can address the challenges of data diversity in specific tasks, it's not appropriate for SD during pre-training. This is because it presupposes knowledge of the evaluation set for downstream tasks, yet it's uncertain which tasks will be targeted by researchers using the pre-trained PLMs in the subsequent stages.

\subsubsection{Ensemble Learning}

EnsLM~\citep{duan-etal-2021-enslm}  proposes an auto-encoding topic model to cluster data, and adopts ensemble learning with weight modulation module to fit different data clusters.
Although it improves cross-domain performance, its data clustering step and weight modulation module are time-consuming and inconvenient to be adopted by PLMs.

\subsubsection{Domain Diversity}

One common type of data diversity is domain diversity. Prior approaches on multi-domain data generally assume that domain labels are well defined and assigned to each sample
~\citep{wright-augenstein-2020-transformer,du-etal-2020-adversarial,jiang-etal-2020-multi-domain}.
To improve the adaptability of the NMT model to various domains, ~\citet{jiang-etal-2020-multi-domain} propose a novel multi-domain NMT model using individual modules for each domain, on which they apply word-level, adaptive and layer-wise domain mixing. 
In order to solve the problem that BERT cannot adapt to domain features for cross domain transfer in cross domain emotion classification tasks, ~\cite{du-etal-2020-adversarial} design a post-training procedure, which contains the target domain masked language model task and a novel similar domain distinguishing task for pre-training.
Although above methods work well in specific downstream domain tasks, they are not suitable for pre-training with source diversity for the following reasons.
First, they generally only design for the multi-domain dataset of a certain task or some specific tasks, rather than the scenario of language model pre-training. 
Second, these pre-defined domain labels are not always accurate or even available~\cite{aharoni-goldberg-2020-unsupervised}, especially for the wild datasets, in which data come from different sources, such as internet news, product reviews, and daily conversations~\cite{duan-etal-2021-enslm}.

\subsection{Multi-task Learning with Prompts}

Prompts have been widely used to better exploit PLMs in downstream tasks under various settings. 
They can be applied during pre-training, fine-tuning and downstream probing.
~\citet{brown2020language} shows that using demonstration examples or instructions as prompts can make GPT-3 accomplish several tasks under few-shot or zero-shot settings.
~\citet{sanh2021multitask} suggests that colossal pre-training corpora contain various task-related data, and appropriate prompting helps PLMs recall such data in downstream applications.
In other words, prompts serve as a bridge between pre-training corpora and downstream tasks.

Therefore, many recent efforts have considered using prompts for multi-task learning. 
Prefix-tuning~\citep{li2021prefix} freezes PLMs and tunes only task-specific prompts for downstream tasks.
T0~\cite{sanh2021multitask} fine-tunes PLMs on a big prompted dataset covering a wide range of tasks, attaining strong zero-shot performance on several tasks.
PPT~\cite{Gu2021PPTPP} generalizes three types of pre-training tasks to learn prompts on big unlabeled datasets, and transfer these prompts to zero-shot task datasets via initialization.
UL2~\cite{tay2022unifying} propose Mixture-of-Denoisers (MoD), a pre-training objective that combines diverse pre-training paradigms together, using a specific prefix for different pre-training methods. They further introduce a notion of mode switching, wherein downstream fine-tuning is associated with specific pre-training schemes. UL2 outperforming T5 and GPT-like models across multiple diverse setups.

The task diversity scenario is close to SD in pre-training. 
The above methods enable PLMs to remember heterogeneous knowledge from various tasks and  retrieve related knowledge in downstream tasks by task-specific prompts.
Inspired by these efforts, we hence prop source prompts to solve the SD problem in pre-training corpora.


\section{Method}

In this section, we describe how source prompts can be applied to enhance language model pre-training with source diversity. 
First, we describe how the source prompt (SP) is implemented in the pre-training stage. 
Then, we describe how the pre-training model with SP is fine-tuned in the downstream task stage.

\subsection{Preliminary}
Pretraining Language Models (PLMs) have resulted in a substantial performance enhancement in Natural Language Processing (NLP) tasks. This improvement is primarily achieved through the use of expansive neural networks pretrained on an extensive corpus of unlabeled data with a self-supervised objective. Contemporary PLMs mainly leverage the transformer architecture~\citep{vaswani2017attention}. There are two principal self-supervised objectives: Masked Language Model (MLM) and Causal Language Model (CLM). The MLM approach involves masking certain tokens within a sentence and predicting them via classification, while CLM predicts the subsequent token based on the token sequence of preceding ones.

In our research, we introduce our method to three different models: encoder-only, encoder-decoder, and decoder-only. The encoder-only and encoder-decoder models are significant components of MLMs, while the decoder-only model is the core element of CLMs.

\subsection{Pre-Training with Source Prompt}
\label{sec:sp-pre}
Towards more powerful PLMs, a common and successful practice is to build more gigantic models and pre-train them on more diverse and colossal corpora. 
A diverse corpus is mainly obtained by converging small ones from multiple sources. 
Therefore, the corpus $\mathcal{C}$ contains multiple subsets from different sources:  
\begin{equation}
    \mathcal{C} = \mathcal{C}_{1}\cup \mathcal{C}_{2}\cup ...\cup \mathcal{C}_{m},
\end{equation}  
where each $ \mathcal{C}_{i} $ is a sub-corpus from a specific source, and $m$ is the number of subsets. 

For example, the corpora of BERT is composed of $m=2$ datasets, Wikipedia and BookCorpus:
\begin{equation}
    \mathcal{C}_{BERT} = Wikipedia \cup BookCorpus.
\end{equation}

A corpus $\mathcal{C}$ (or sub-corpus $\mathcal{C}_i$) is a set of $n$ pieces of texts: 
\begin{equation}
    \mathcal{C} = \{ t_{1}, t_{2}, ..., t_{n} \}.
\end{equation}
We assign each sub-corpus $\mathcal{C}_i$ a name $N_i$ indicating its source. 
Then, for every $t_{j} \in \mathcal{C}_{i}$, we define $\mathcal{C}_i$ as the \textit{source corpus} (or \textit{source}) of $t_j$ and $N_i$ as its \textit{source name}. 
A simple yet effective way of naming these sources is to use their abbreviation, namely \textbf{abbreviation SP}, such as ‘[WIKI]’ for Wikipedia and ‘[BOOK]’ for BookCorpus. 
In fact, our method is \textbf{highly robust to different naming methods}, and it is  even feasible to use meaningless letters to denote the sources instead, such as A, B and C.

\begin{figure*}[!htb]
    \centering
    \includegraphics[width=1.9\columnwidth]{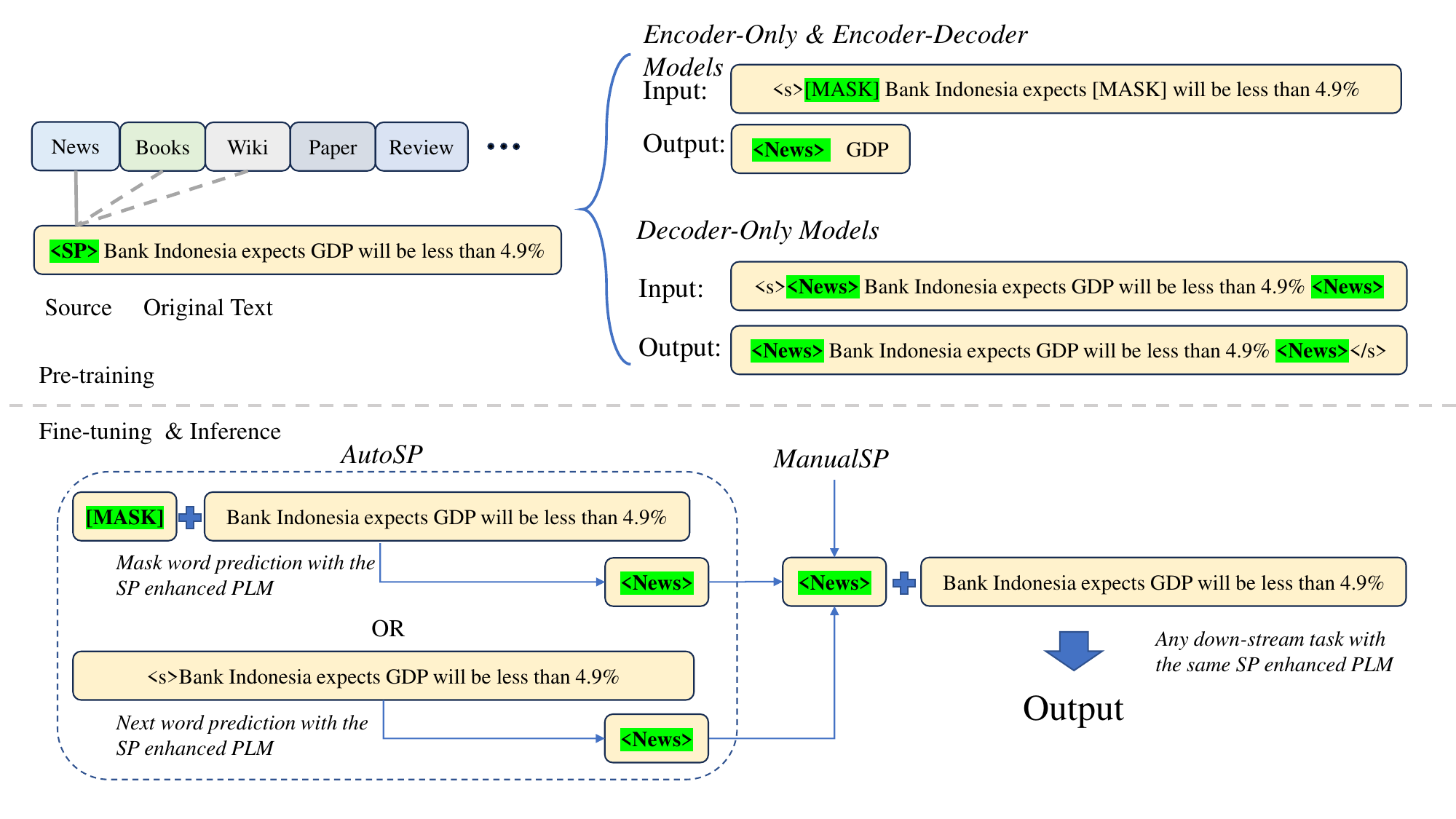}
    \caption{For encoder-only or encoder-decoder models, we employ Masked Source Prediction task, training it with Masked Language Modeling. For decoder-only models, we utilize Next Token Prediction in pre-training. During fine-tuning and inference, SP can be manually added for data from known sources (manual SP), or automatically appended MSP's or NTP's prediction of the closest source (auto SP) for data from unknown sources. we can also fine-tune and inference without any SP.
    }
    \label{fig:sp-pt-ft}
\end{figure*}

\subsubsection{SP Pre-training on Encoder-Only or Encoder-Decoder Models}
We place a source prompt at the start of each text during pre-training, thereby informing the PLMs of the text's origin. For every $t_{j}$ in $\mathcal{C}_{i}$, its SP corresponds to the corpus name $N_i$. We separate the source prompt and the original input using a distinct delimiter. Therefore, the source-prompted input $\hat{t}_j$ becomes:
\begin{equation}
\hat{t}_j = [N_i;-;t_j],
\end{equation}
where $[\cdot;\cdot]$ denotes string concatenation. For instance, considering the following text extracted from a news corpus:

``\textit{Spokesman for the British Prime Minister Johnson: we hope to make significant amendments to the Northern Ireland agreement. We believe it is feasible within the agreement's framework. }''

We prefix this text with its source name, \textit{News}:

``\textit{News} $-$ \textit{Spokesman for the British Prime Minister Johnson: we hope to make significant amendments to the Northern Ireland agreement. We believe it is feasible within the agreement's framework}''

Subsequently, this source-prompted text is subjected to the pre-training phase of the PLMs. Therefore, the PLMs can predict masked words using the SP's assistance, enabling them to learn language styles dependent on varying sources. Simultaneously, the tokens' representations in the SP are optimised, leading to an incremented source-specific knowledge for both pre-training and downstream applications.
Furthermore, we introduce the masked source prediction (MSP), an innovative pre-training objective for multi-source pre-training. In this process, the source prompts are randomly masked with a certain probability, requiring the PLMs to predict the masked source based on the contexts. This task compels PLMs to learn-source related characteristics. It can be easily amalgamated with the MLM objective and incurs minimal additional costs.

\subsubsection{SP Pre-training on Decoder-Only Models}
The decoder-only model, lacking a bidirectional attention mechanism, cannot apply MLM. We propose two types of SPs for such models: 
\begin{enumerate}
    \item  \textbf{SP} Positioning the SP at the start of the text allows the SP to assist in predicting the text's content.
    \item  \textbf{Post SP} Placing the SP at the text's end enables source prediction based on the text's content.
\end{enumerate}
We use the next-token-prediction method for training these models.

It's necessary to emphasize that during the pre-training phase of any architectures, SP is incorporated at random, and a special token is interspersed between the SP and the original text. 

\subsection{Fine-tuning with Source Prompts}
\label{sec:finet}

Applying PLMs with SP to a specific downstream dataset involves assigning the most relevant pre-training source prompt to the dataset. Certain datasets derive from specific sources or domains closely related to one of the pre-training sources, for example, a news classification dataset. Therefore, a manual selection of an SP for such datasets is feasible. Conversely, other datasets might have less relevance to pre-training sources or may include samples from various sources. In these cases, we suggest two methodologies for assigning source prompts to downstream datasets:

\subsubsection{Manual SP} Often, we compile downstream datasets from a single source, detailed in its description. Given this scenario, we can manually assign it an appropriate pre-training source, if available. For instance, when pre-training PLMs on Wikipedia and BookCorpus, similar to BERT, and fine-tuning models on the QNLI~\cite{wang2018glue} dataset (a dataset precisely derived from Wikipedia), we can directly employ the SP [WIKI] for it. Following SP selection, we insert the SP before the texts of the downstream datasets, aligning with the pre-training process.

\subsubsection{Auto SP}
Alternatively, downstream dataset sources could be unidentified, mixed, or significantly divergent from the pre-training sources. For these situations, we suggest the Auto SP methodology, as demonstrated in Figure~\ref{fig:sp-pt-ft}. The Auto SP method leverages our MSP objective during pre-training, enabling MLMs to self-predict the most suitable sources. Specifically, we initially input the concatenation $[$[MASK]$;-;t]$ of the source mask and the original input $t$. We then ask PLMs to predict each sample's text source, including both training and testing data. Or predict the source with Next Token Prediction for CLMs. Finally, we replace the masked source (MASK) with the predicted source, using these substituted samples for fine-tuning and prediction.

\subsubsection{No SP} Additionally, it is possible to use the pre-trained model conventionally without any SP. Firstly, the application of SP during pre-training enables the model to comprehend the source corpus effectively. Secondly, the optional random addition of SP in the settings imparts a degree of generalizability to the model, irrespective of the presence of a SP. Consequently, even in the absence of SP implementation in downstream tasks, comparable performance levels can be achieved as with SP.


\section{Experiments}

In this section, we evaluate the effectiveness of our source prompt method under different settings. In general, PLMs pre-trained with our method on diverse corpora achieve significant improvements.

We begin by outlining the consistent settings employed across all experiments. Subsequently, we undertake the following experimental tasks:
\begin{enumerate*}[label=\alph*).]
    \item Verification of the effectiveness of our proposed method.
    \item Comparison of the impacts of diverse naming policies for pre-training corpus sources, demonstrating the robustness of SP to these varying policies.
    \item Analysis of different masking probabilities for SP, attesting to the effectiveness of MSP.
    \item Evaluation of the effects of various SP assignment methods for downstream datasets, revealing that auto SP is the most efficacious approach.
    \item Confirmation of the generalizability of our methodology across different model architectures, pre-training corpora and downstream benchmarks.
    \item Investigation of the effects of SP methods on domain-specific or general corpora for decoder-only architecture large language models.
\end{enumerate*}

\subsection{General Settings}
\label{sec:settings}
\subsubsection{Corpus}

We consider three diverse corpora of multiple sources for pre-training, including BBT-FinCorpus\footnote{https://github.com/ssymmetry/BBT-FinCUGE-Application}, CLUECorpusSmall~\citep{CLUECorpus2020} and WuDaoCorpora~\citep{yuan2021wudaocorpora}, which represent typical domain-specific corpora or general corpora, respectively. Detailed information of curpua is shown in Appendix.

\begin{table*}[!htb]
\centering
\resizebox{2.1\columnwidth}{!}{
	\begin{tabular}{l l l c c c c c c c c c}
	\hline
	\textbf{Setting} & \textbf{Pre-train} & \textbf{Fine-tune} & \textbf{FinCQA} & \textbf{FinESE} & \textbf{FinFE} & \textbf{FinNA} & \textbf{FinNL} & \textbf{FinNSP} & \textbf{FinQA} & \textbf{FinRE} & \textbf{Avg. }\\
	\hline
	A & w/o SP & w/o SP  & 67.81 & 78.84 & 79.85 & 42.37 & 87.28 & 89.13 & 74.75 & 54.08 & 71.76 \\
	B & Abbreviation SP & w/o SP & 77.14 & 78.89 & 78.26 & 46.15 & 87.75 & 90.56 & 81.90 & 56.68 & 74.67 \\
	C & Abbreviation SP & Manual SP & \textbf{77.75} & 79.25 & 78.96 & 46.47 & 87.82 & 90.56 & 81.76 & \textbf{57.19} & 74.97 \\
	D & Abbreviation SP & Auto SP & 76.99 & 78.89 & 79.75 & \textbf{56.64} & \textbf{87.93} & 90.56 & 81.04 & 56.12 & \textbf{75.99} \\
	E & Abbreviation SP & Random SP & 77.31 & \textbf{79.84} & 79.35 & 50.63 & 87.76 & \textbf{93.86} & \textbf{83.32} & 54.99 & 75.88\\
	\hline
	\end{tabular}
}
\caption{Results of the T5 model pre-trained with BBT-FinCorpus and fine-tuned with BBT-FinCUGE. In general, T5 with SP performs significantly better than  T5 without SP. As shown in rows C, D, and E, auto SP outperforms the other assignment method for downstream datasets.}
\label{table:fint5}
\end{table*}

\begin{table*}[!htb]
\centering
\resizebox{2.1\columnwidth}{!}{
	\begin{tabular}{l l c c c c c c c c c}
	\hline
	\textbf{Setting} & \textbf{SP mask prob.} & \textbf{FinCQA} & \textbf{FinESE} & \textbf{FinFE} & \textbf{FinNA} & \textbf{FinNL} & \textbf{FinNSP} & \textbf{FinQA} & \textbf{FinRE} & \textbf{Avg. }\\
	\hline
        A & 0 & 76.10 & 78.90 & 78.56 & 45.63 & 87.66 & 89.80 & 80.54 & 55.23 & 74.05 \\
	B & 0.15 & 76.99 & 78.89 & \textbf{79.75} & \textbf{56.64} & 87.93 & 90.56 & 81.04 & 56.12 & 75.99 \\
        C & 0.3 & \textbf{77.18} & \textbf{79.20} & \textbf{79.75} & \textbf{56.64} & \textbf{88.05} & \textbf{90.98} & \textbf{81.42} & \textbf{56.35} & \textbf{76.19} \\
	\hline
	\end{tabular}
}
\caption{Experiments with pre-trained T5 with or without masked source prediction (MSP) objective. The results show the effectiveness of MSP.}
\label{table:msp}
\end{table*}

\subsubsection{Benchmark}

We use BBT-FinCUGE and CLUE~\cite{xu2020clue} as our evaluation benchmarks. 
BBT-FinCUGE is a Chinese financial evaluation benchmark consisting of five understanding tasks and three generation tasks. The understanding tasks include event subject extraction, emotion recognition, news classification, negative news and subject recognition, and relationship extraction. The generation tasks include causal QA, event QA and news summary. 
CLUE is a general Chinese NLP evaluation benchmark consisting of nine comprehension tasks, including semantic similarity, text classification, reading comprehension and other tasks.
For all evaluation benchmarks, we fine-tune PLMs for evaluation, and take the average score on the test set as the main comparison basis.

\subsubsection{Implementation}

We delineate the specifics of our pre-training and fine-tuning stages in this section.

Our selected foundational architectures encompass two conventional Masked Language Models (MLMs), BERT and T5, as well as one of the most recent Causal Language Models (CLMs), OpenLLaMA-3b. BERT exemplifies the Encoder-Only Model, T5 embodies the Encoder-Decoder Model, whereas OpenLLaMA-3b signifies the Decoder-Only Model. Our implementation of these models is heavily reliant upon Hugging Face Transformers. The configuration parameters for BERT-base and T5-base align with their original implementations. For OpenLLaMA-3b, we incorporate an expanded vocabulary to encompass Chinese tokens, as suggested by previous research.

We execute pre-training for the three models on either domain-specific corpora or general corpora, and subsequently assess their performance on benchmarks such as BBT-FinCUGE or CLUE. A comprehensive depiction of the implementation is provided in the Appendix. To ensure the reliability of our findings, all experiments are conducted thrice and the average results are reported.

\subsection{Overall Effectiveness of SP}
\label{sec:sp-effect}



To verify the basic effectiveness of SP method under the pre-training and fine-tuning framework, we pre-train T5 on BBT-FinCorpus and fine-tune it on the BBT-FinCUGE benchmark. We set up four experimental groups: group A, which does not use SP in both stages, group B, which uses SP only in the pre-training stage, and groups C and D, which use SP in both stages and adopt manual SP and auto SP respectively in the fine-tuning stage.

Table~\ref{table:fint5} shows the results, from which we make the following observations:
(1) With source prompts, PLMs pre-trained on diverse corpora achieve significantly better performance on nearly all datasets. Their average scores (74.67-75.88) have significant advantages over group A (71.76), which fully proves the effectiveness of the SP method, especially with nearly no additional computing cost.
(2) Introducing SP in the fine-tuning phase further improves model performance, as shown by comparing group B with groups C, D, and E.

\begin{table*}[!htb]
\centering
\resizebox{1.8\columnwidth}{!}{
	\begin{tabular}{l l l c c c c c c}
	\hline
	\textbf{Setting} & \textbf{Pre-train} & \textbf{Fine-tune} & \textbf{FinESE} & \textbf{FinFE} & \textbf{FinNL} & \textbf{FinNSP} & \textbf{FinRE} & \textbf{Avg. }\\
	\hline
	A & w/o SP & w/o SP & 38.44 & 77.22 & 83.77 & 68.19 & 44.44 & 62.42 \\
	B & Abbreviation SP & w/o SP & 37.37 & 77.17 & 83.06 & 68.20 & 45.71 & 62.30 \\
	C & Abbreviation SP & Manual SP & 58.45 & \textbf{77.82} & 83.76 & 67.62 & 49.14 & 67.36 \\
	D & Abbreviation SP & Auto SP & 57.34 & 77.77 & 82.72 & 69.82 & \textbf{50.50} & 67.53 \\
	E & Alphabet SP & Manual SP & 56.21 & 77.47 & 83.60 & 68.19 & 45.70 & 66.23 \\
	F & Alphabet SP & Auto SP & \textbf{59.37} & 77.07 & 83.22 & \textbf{70.18} & 45.15 & 67.00 \\
	G & Misplaced SP & Manual SP & 57.49 & 77.57 & \textbf{84.24} & 69.84 & 49.02 & \textbf{67.63} \\
	H & Misplaced SP & Auto SP & 57.77 & 77.73 & 84.10 & 69.65 & 47.77 & 67.32 \\
	\hline
	\end{tabular}
}
\caption{Experiments of BERT in financial domain. It is pre-trained on BBT-FinCorpus and fine-tuned on BBT-FinCUGE. As shown in row C to H, SP is robust to different source naming policies in the pre-training stage.}
\label{table:finbert}
\end{table*}


\subsection{Robustness of Source Naming Policies}
\label{sec:namingsource}

As mentioned in Method Section, we do not have a deterministic method to name the source of each corpus. 
A relatively simple way is to use the manual abbreviation of the corpus source. However, we show with experiments that our method is robust to different source naming policies, namely the specific tokens used to represent the sources.
That is to say, the effectiveness of SP originates from identification of sources, instead of textual information in their names. 

Specifically, we design two additional experiments for comparison. 
The first is meaningless alphabet SP, that is, we replace the specific names of the corpus sources with meaningless letters like A, B, C, etc. 
Thus, the model can only obtain the source identification of corpora, without specific textual information of each source. 
The second is misplaced SP, that is, we deliberately confuse the names of the corpus sources (for example, set the SP of all news corpora as ``comments''). 
In all settings, we control the number of prompt tokens of different sources to be equal.

Table~\ref{table:finbert} shows the results on BERT. The results demonstrate that using alphabet SP and misplaced SP hardly decrease the performance, compared with abbreviation SP, which is still far beyond the baseline without SP. These results suggest that SP is robust to the different name policies of corpus. 
Hence, SP is effective because of identification of sources, instead of textual information in their names.  
Therefore, in real applications, the sources can be named at will, which exert little influence on the performance.

\begin{table*}[!htb]
\centering
\resizebox{2\columnwidth}{!}{
    \begin{tabular}{l l l c c c c c c c}
    \hline
        \textbf{Setting} & \textbf{Pre-Train} & \textbf{Fine-Tune} & \textbf{AFQMC} & \textbf{CSL} & \textbf{IFLYTEK} & \textbf{OMNLI} & \textbf{TNEWS} & \textbf{WSC} & \textbf{Avg.}\\
        \hline
        A & w/o SP & w/o SP & 69.45 & 77.53 & 57.57 & 75.88 & 55.31 & 63.71 & 66.57 \\
        B & Abbreviation SP & w/o SP & 69.11 & 77.36 & \textbf{58.06} & 75.64 & 55.20 & \textbf{63.81} & 66.53 \\
        C & Abbreviation SP & Auto SP & \textbf{70.92} & 78.13 & 57.82 & 75.59 & \textbf{55.44} & \textbf{63.81} & 66.95 \\
        D & Abbreviation SP & Manual SP & \textbf{70.92} & \textbf{79.36} & \textbf{58.06} & \textbf{76.32} & 55.51 & \textbf{63.81} & \textbf{67.33} \\
    \hline
    \end{tabular}
}
\caption{Experiments of BERT in the general domain. It is pre-trained on the CLUECorpusSmall and fine-tuned on the CLUE benchmark. The results prove the effectiveness of SP in the general domain, and show the generalizability of SP.}
\label{table:cluebert}
\end{table*}

\begin{table*}[!htb]
\centering
\resizebox{2.1\columnwidth}{!}{
	\begin{tabular}{l l l c c c c c c c c c}
	\hline
	\textbf{Setting} & \textbf{Pre-train} & \textbf{Fine-tune} & \textbf{FinCQA} & \textbf{FinESE} & \textbf{FinFE} & \textbf{FinNA} & \textbf{FinNL} & \textbf{FinNSP} & \textbf{FinQA} & \textbf{FinRE} & \textbf{Avg. }\\
	\hline
	A & w/o SP & w/o SP  & 75.4 & 84.7 & 79.8 & 48.7 & 88.0 & 94.8 & 79.9 & 55.6 & 75.8 \\
	B & Abbreviation SP & w/o SP & 75.7 & 84.3 & \textbf{80.1} & 49.7 & 87.8 & \textbf{96.0} & 80.6 & 56.6 & 76.4 \\
	C & Alphabet SP & w/o SP & 75.6 & \textbf{86.1} & 78.4 & 49.5 & 87.7 & 95.4 & \textbf{81.5} & 55.3 & 76.2 \\
	D & Post Abbreviation SP & w/o SP & \textbf{77.3} & 85.2 & 80.0 & \textbf{49.9} & \textbf{88.3} & 95.6 & 81.0 & 55.5 & \textbf{76.6} \\
	\hline
	\end{tabular}
}
\caption{Results of the OpenLLaMA-3b model pre-trained with BBT-FinCorpus and fine-tuned with BBT-FinCUGE. All pre-trained models are fine-tuned without SP. As shown in rows B, C, and D, all SP implenmentation can improve model generally, and Post Abbreviation SP achieves the best.}
\label{table:finopenllama}
\end{table*}

\begin{table*}[!htb]
\centering
\resizebox{2\columnwidth}{!}{
    \begin{tabular}{l l l c c c c c c c}
    \hline
        \textbf{Setting} & \textbf{Pre-Train} & \textbf{Fine-Tune} & \textbf{AFQMC} & \textbf{CSL} & \textbf{IFLYTEK} & \textbf{OCNLI} & \textbf{TNEWS} & \textbf{WSC} & \textbf{Avg.}\\
        \hline
        A & w/o SP & w/o SP & 71.95 & 84.93 & 60.38 & \textbf{77.67} & 47.59 & \textbf{68.57} & 68.57 \\
        B & Abbreviation SP & w/o SP & 72.86 & 85.57 & \textbf{62.77} & 77.37 & 59.68 & 64.8 & \textbf{70.5} \\
        C & Post Abbreviation SP & w/o SP & \textbf{73.22} & \textbf{85.67} & 61.19 & 77.07 & \textbf{60.38} & 62.41 & 69.9 \\
    \hline
    \end{tabular}
}
\caption{Experiments of OpenLLaMA-3b in the general domain. It is pre-trained on the CLUECorpusSmall and Wudao Corpus, and fine-tuned on the CLUE benchmark without SP. The results prove the effectiveness of SP in the general domain with scaled model and training tokens.}
\label{table:clueopenllama}
\end{table*}

\subsection{Effectiveness of Mask Source Prediction}
\label{sec:exp-msp}

We study the effectiveness of the mask source prediction (MSP) objective by pre-training models with varying mask probability of SP and comparing their performance on the benchmark.

We consider three  values \{0, 0.15, 0.3\}  of the mask probability $P$.
Setting $P=0$ means that the models are pre-trained without the MSP objective. 
We use abbreviation SP for pre-training and  manual SP for fine-tuning. 
As shown in the table~\ref{table:msp}, the performance with $P=0$ (without MSP) is significantly under that of others, 
and the performance with $P=0.3$ is slightly better than that with $P=0.15$.
These validate the effectiveness of our MSP objective, and suggest that higher  SP masking probability encourage the model to better distinguish texts from different sources and learn source related features.

\subsection{SP Assignment for Downstream Tasks}
\label{sec:exp-ft}



As described, we propose two methods to assign SP in downstream datasets: manual SP and auto SP. This experiment compares the effects of different SP assignment methods. We also set up a control group with a randomly sampled SP from the pre-training sources for each sample, called random SP.
Table \ref{table:fint5} shows that auto SP outperforms manual SP (rows C and D). We attribute this to the fact that auto SP leverages the source related information learned by the model, and adaptively applies the most suitable SP for each sample, while manual SP is only assigned at the dataset-level. This conclusion is further validated by rows C and D in Table~\ref{table:finbert} and Table \ref{table:cluebert}.

\subsection{Generalizability of SP}
\label{sec:exp-gen}

In order to verify the generalization of SP method, we mainly conduct experiments from two dimensions, including different model architectures and different corpus domains. 
Specifically, in order to verify the generalization of SP to different model architectures, we applied SP methods to BERT and T5 models respectively. 
In order to verify generalization of SP to different domains, we conducted experiments in both the financial domain and the general domain. 
The corpus and benchmark used are BBT-FinCorpus and BBT-FinCUGE in the financial domain, and are CLUECorpusSmall and CLUE in the general domain.

Table~\ref{table:fint5} and Table~\ref{table:finbert} show that our models achieve significant improvement under both model architectures. 
Besides, it is shown in Table~\ref{table:finbert} and Table~\ref{table:cluebert} that our model has the expected effect in both domains. 
Therefore, it can be concluded that our method is general enough to be widely applied to various model architectures and corpus domains. 

\subsection{SP on Decoder-Only LLMs}
We conducted systematic experiments on the OpenLLaMA-3b model to examine the influence of SP on decoder-only LLM. In order to explore the effects of the two SP methods, we conduct experiments on two SP separately.
Table \ref{table:finopenllama} and Table \ref{table:clueopenllama} depict the experimental outcomes of OpenLLaMA-3b in the financial and general domains, respectively. All SP settings achieved scores beyond the preset baseline in both experiments, implying the enhancement in model performance during the pre-training process due to the incorporation of SP. Moreover, all downstream fine-tuning procedures were executed without SP, thus highlighting the robustness of the SP.

\section{Conclusion}
 
In this paper, we first identify the side-effects of increased corpus diversity for pre-training PLMs.
To overcome this problem, we propose source prompt (SP), an easy, efficient and effective approach to promote coordinated pre-training on such corpora, 
which is a prompt added before inputs of PLMs to identify their source. 
Furthermore, we thoroughly study different naming polices of pre-training SP and different strategy to assign SP to downstream application, as well as proposing a novel pre-training objective named masked source prediction.
Results of extensive experiments validate the effectiveness, robustness and generalizability of SP, as well as the benefits of MSP. 

\section*{Limitations}
First of all, "source" is a relatively abstract concept. For common crawl based corpora such as C4, their source information is largely unusable because their data is crawled from millions of web pages. Therefore, our current method is limited to the scenario where a certain number of small corpora are merged together to form a large corpus.
Second, due to the scale of computing power we can obtain, the parameter scale and the amount of training tokens of PLMs we use are quite limited. 
It remains to be studied whether our method works for the large-scale PLMs.
Last but not least, it remains to be explored in-depth what the effectiveness of introducing SP originates from.
In the future, we will continue to study the effect of SP on large-scale PLMs, and investigate the effectiveness of SP in other NLP tasks such as cross-domain sentiment analysis.

\bibliography{aaai24}

\section{Appendix}
\subsection{Corpora Details}
BBT-FinCorpus represents a substantial Chinese financial corpus comprising approximately 200GB of text files. This corpus incorporates various sources such as company announcements, research reports from securities companies and investment banks, discussion forums focused on stock bars and applicable forums such as the Snowball forum, as well as multiple financial news fetched from several websites. As demonstrated in Table~\ref{tab:diffcorpus}, these five distinct sources propose a challenge for models attempting to learn due to the differentiation in their styles.

\begin{table*}[!htb]
\centering
\resizebox{2.1\columnwidth}{!}{
\begin{tabular}{p{4cm} p{9.5cm} p{8cm}}
\hline
\textbf{Corpus} & \textbf{Example} &\textbf{Description}\\
\hline
Company Announcement
& \textit{The 2021 annual equity distribution plan of Changying Technology Co., Ltd. }
& Announcements of listed companies, formal
\\
Research Report
& \textit{After the epidemic, macro-economy will recover after the trough...}
& Research reports about companies, industry and economy
\\
Guba BBS
& \textit{I hope medical stocks will improve tomorrow, buy in!}
& Discussion of shareholders, colloquial and emotional
\\
Snowball BBS
& \textit{I think Bitcoin will rise due to the release of the US dollar. }
& Discussion of shareholders, colloquial and rational
\\
Financial News
& \textit{Musk responded that Tesla stopped receiving orders for ...}
& Financial news, diverse and relatively formal
\\
\hline
\end{tabular}}
\caption{Sources, examples and descriptions of various sub-corpora in the BBT-FinCorpus.}
\label{tab:diffcorpus}
\end{table*}

CLUECorpusSmall signifies a general Chinese corpus possessing about 14GB of text files. The corpus is composed of four diverse sources, each demonstrating a significant difference in their styles.

WudaoCorpora is a representation of a general Chinese corpus that contains approximately 220GB of text files. The corpus is made up of 25 data types where each type indicates a source from which the data has been selected.

A comprehensive list of the data sources for these three corpora is exhibited in Table \ref{table:corporasources}.

\subsection{Implementation Details}
DeepSpeed~\cite{rasley2020deepspeed} is employed to expedite the training process. Notably, we adopt the BFLOAT16~\cite{kalamkar2019study} semi-precision format and gradient partitioning, as implemented in DeepSpeed. For OpenLLaMA-3b's pre-training, to accelerate the computational process, we utilize flash-attention ~\cite{dao2023flashattention2}.

BERT's pre-training was executed using a batch size of 128, sequence length of 512, a learning rate of 5e-5, and a total of 100,000 steps. This took approximately 24 hours using 8 NVIDIA A100 GPUs. For evaluation, BBT-FinCUGE's three generation tasks were omitted and a simple fully-connected layer served as the output head for each task, based on BERT's last hidden state.

T5's training strategy on BBT-FinCorpus adhered to UER~\cite{zhao2019uer}'s two-stage setting. In the initial stage, the model was trained using a sequence length of 128, a batch size of 512, and a total of 1,000,000 steps. During the second stage, with a sequence length of 512, a learning rate of 1e-4, a batch size of 128, and 250,000 total steps, the model was trained within 60 hours using 8 NVIDIA A100 GPUs. For general domain, due to the scale limitation of CLUECorpusSmall, the training strategy is to train 100,000 total steps on it under the same settings. Evaluations modeled all tasks as text-to-text. Unnecessary variables were avoided by not using task prompts; the model was trained to output the relationship between head entities and tail entities in text, based on input sentences, head entities, and tail entities.

OpenLLaMA-3b was trained using batch sizes of 512, a sequence length of 1024, learning rates of 5e-5, and 20,000 total steps, taking around 70 hours with 8 NVIDIA A800 GPUs for both general domain and finicial domain. Owing to the scale limitation of CLUECorpusSmall, WudaoCorpora was included in the general domain pre-training. The exact numbers of training tokens are presented in table \ref{table:generaldomaintokens}. For evaluation, input and target were concatenated to model tasks as text-to-text. During loss computation, only the MSE loss on the target token positions was considered.

For Fine-tuning, we train with batch size of 64 and learning rate of 5e-5 on the T5 and Bert models. While on OpenLLaMA-3b, we train it with batch size of 192 and learning rate of 1e-5. We use the same parameters for Fine-tuning on BBTFinCUGE and CLUE~\footnote{https://github.com/CLUEbenchmark/CLUE}. To ensure sufficient fine-tuning, we test on the testset with the model that scores the highest on the validation set after 8 training rounds on the downstream tasks.

\begin{table}[h]
\small
\centering
    \begin{tabular}{l c}
    \hline
        \textbf{Corpus} & \textbf{Sources} \\
        \hline
        \multirow{25}*{WudaoCorpora} & Douban Topic \\
         & Blog \\
         & Nurturing Common Sense \\
         & Medical Question and Answering \\
         & Science and Technology \\
         & Introduction to Xiaohongshu \\
         & Agriculture \\
         & Encyclopedia \\
         & Entertainment \\
         & Information \\
         & Economy \\
         & Baijiahao Article \\
         & Culture \\
         & News \\
         & Sociaty \\
         & Experience \\
         & Travel \\
         & Real Estate \\
         & Education \\
         & International \\
         & Games \\
         & Sports \\
         & Cars \\
         & Law \\
         & Popular Science Articles \\
        \hline
        \multirow{4}*{CLUECorpusSmall} & comments\\
        & news \\
        & webtext \\
        & wikizh \\
        \hline
        \multirow{5}*{BBT-FinCorpus} & Company Announcement \\
        & Research Reports \\
        & Guba BBS \\
        & Snowball BBS \\
        & Financial News \\
    \hline
    \end{tabular}
\caption{The detailed list of data source from WudaoCorpora, CLUECorpusSmall and BBT-FinCorpus}
\label{table:corporasources}
\end{table}

\begin{table}[h]
\small
\centering
    \begin{tabular}{l c}
    \hline
        \textbf{Corpus} & \textbf{Tokens} \\
        \hline
        WudaoCorpora & 6 Billion \\
        CLUECorpusSmall & 4 Billion\\
    \hline
    \end{tabular}
    \caption{The number of training tokens from WudaoCorpora and CLUECorpusSmall. The number of tokens from full CLUECorpusSmall is 4 billion.}
\label{table:generaldomaintokens}
\end{table}

\end{document}